\documentclass[a4paper]{article}
\usepackage{etoolbox}

\usepackage{amsmath,amsthm}
\usepackage{mathtools}
\usepackage[dvipsnames,table]{xcolor}

\usepackage{natbib}
\usepackage{authblk}

\RequirePackage{crop}
\RequirePackage{graphicx}
\usepackage[labelfont=it]{caption}
\RequirePackage{amsmath}
\RequirePackage{array}
\RequirePackage{color}
\RequirePackage{xcolor}
\RequirePackage{amssymb}
\RequirePackage{flushend}
\RequirePackage{stfloats}
\RequirePackage[figuresright]{rotating}
\RequirePackage{chngpage}
\RequirePackage{totcount}
\RequirePackage{fix-cm}
\RequirePackage{hyperref}
\usepackage{doi}

\definecolor{linkcolor}{HTML}{991408}  
\definecolor{citecolor}{HTML}{2E7E2A}  
\definecolor{filecolor}{HTML}{131877}  
\definecolor{menucolor}{HTML}{727500}  
\definecolor{runcolor} {HTML}{137776}  
\definecolor{urlcolor} {HTML}{0a2bbf}  
\hypersetup{colorlinks=true,linkcolor=linkcolor,citecolor=citecolor,filecolor=filecolor,menucolor=menucolor,runcolor=runcolor,urlcolor=urlcolor}
\newtoggle{arxiv}
\toggletrue{arxiv}
\iftoggle{arxiv}{
  \usepackage[parfill]{parskip}

  \usepackage[%
    a4paper,
    inner=25mm,
    outer=25mm,
    top=25mm,
    bottom=25mm,
    marginparsep=5mm,
    marginparwidth=40mm,
  ]{geometry}

  \newlength{\defbaselineskip}
  \setlength{\defbaselineskip}{\baselineskip}
  \setlength{\parskip}{6pt}%
  \setlength{\parindent}{0pt}%

  \RequirePackage[T1]{fontenc}
  \RequirePackage[tt=false, type1=true]{libertine}
  \RequirePackage[varqu]{zi4}
  \RequirePackage[libertine]{newtxmath}
}{}

\usepackage{booktabs}
\usepackage{makecell}
\usepackage{placeins}

\usepackage{multirow}

\usepackage{tikz}
\usepackage{float}
\usepackage{subcaption}

\usepackage{amssymb}
\usepackage{pifont}

\graphicspath{{Figures/}}

\usepackage{comment}

\newcommand{\na}[1]{{\color{darkgray}{N/A}}}

\newcommand{\appref}[1]{\hyperref[#1]{Appendix~\ref*{#1}}}

\definecolor{ForestGreen}{RGB}{34,139,34}

\title{Enhancing DNA Foundation Models to Address Masking Inefficiencies}

\usepackage{fancyhdr}
\usepackage{multicol}
\makeatletter
\renewcommand\@author{
    \AB@authlist\\[\affilsep]
    \begin{minipage}{0.9\textwidth}
    \begin{multicols}{2}
        \raggedright
        \AB@affillist
    \end{multicols}
    \end{minipage}
    }
\makeatother

\author[1,$\ast$]{Monireh~Safari}
\author[1,$\ast$]{Pablo~Millan~Arias}
\author[4,]{\\Scott~C.~Lowe}
\author[1,$\sharp$]{Lila~Kari}
\author[3,5]{Angel~X.~Chang}
\author[2,4,$\sharp$]{Graham~W.~Taylor}

\affil[1]{University of Waterloo}
\affil[2]{University of Guelph}
\affil[3]{Simon Fraser University}
\affil[4]{Vector Institute}
\affil[5]{Alberta Machine Intelligence Institute (Amii)}

\affil[$\ast$]{Joint first author}{ }

\affil[$\sharp$]{Corresponding authors: \href{mailto:gwtaylor@uguelph.ca}{gwtaylor@uguelph.ca}, \phantom{$\sharp$}\href{mailto:lila@uwaterloo.ca}{lila@uwaterloo.ca}}

\date{}

\usepackage{pifont}%

\newcommand{\best}[1]{\textbf{#1}}
\newcommand{\secbest}[1]{\underline{#1}}

\begin{document}

\maketitle

\begin{abstract}

\noindent
Masked language modelling (MLM) as a pretraining objective has been widely adopted in genomic sequence modelling. While pretrained models can successfully serve as encoders for various downstream tasks, the distribution shift between pretraining and inference detrimentally impacts performance, as the pretraining task is to map \texttt{[MASK]} tokens to predictions, yet the \texttt{[MASK]} is absent during downstream applications. This means the encoder does not prioritize its encodings of non-\texttt{[MASK]} tokens, and expends parameters and compute on work only relevant to the MLM task, despite this being irrelevant at deployment time.
In this work, we propose a modified encoder-decoder architecture based on the masked autoencoder framework, designed to address this inefficiency within a BERT-based transformer. We empirically show that the resulting mismatch is particularly detrimental in genomic pipelines where models are often used for feature extraction without fine-tuning. We evaluate our approach on the BIOSCAN-5M dataset, comprising over 2 million unique DNA barcodes. We achieve substantial performance gains in both closed-world and open-world classification tasks when compared against causal models and bidirectional architectures pretrained with MLM tasks. 
\end{abstract}
\section{Introduction}

DNA foundation models have emerged as effective tools for analyzing genomic sequences, utilizing a wide variety of architectures, including transformers \citep{Zhou2021DNABERT,arias2023barcodebert, zhou2024dnaberts}, state space models (SSMs) \citep{poli2023hyena, gao2024barcodemambastatespacemodels}, and convolutional neural networks (CNNs) \citep{Banegas_2023}. These models leverage different pretraining strategies, from causal to bidirectional learning, enabling strong performance across diverse genomic tasks. Among these pretraining strategies, masked language modelling (MLM) has become widely adopted, enabling models to learn effective sequence representations for downstream tasks like specimen identification to taxon and species discovery. However, the effectiveness of MLM is highly dependent on how masking is implemented, as different strategies can affect the model's performance.

In DNA sequence modelling, foundation models typically adopt BERT's three-part masking strategy \citep{Devlin2019BERTPO}, where 80\% of selected tokens are replaced with \texttt{[MASK]}, 10\% remain unchanged, and 10\% are randomly substituted. Models such as the Nucleotide Transformer \citep{NucleotideTransformerDalla-Torre2023} and BarcodeBERT \citep{arias2023barcodebert} followed this approach, while DNABERT \citep{Zhou2021DNABERT} and DNABERT-2 \citep{zhou2023dnabert2} adopted a simpler strategy, replacing 100\% of selected positions with \texttt{[MASK]} tokens. Despite its popularity, MLM introduces a notable limitation: a distribution shift between pretraining and inference due to the absence of \texttt{[MASK]} tokens during downstream tasks. This mismatch leads to representational inefficiencies, as models prioritize the quality of encodings and predictions corresponding to \texttt{[MASK]} tokens but lack a direct target for non-\texttt{[MASK]} token inputs. Consequently, they allocate parameters and compute to tokens never encountered during inference, potentially limiting their ability to capture biologically relevant patterns. While this limitation and its impact on model performance have been studied in natural language processing (NLP) settings \citep{meng2024maelm, Electra}, its effects on DNA sequence foundation models remain unexplored.

In this study, we propose ‌BarcodeMAE which uses a modified encoder-decoder architecture based on the masked autoencoder for MLM (MAE-LM; \citealp{meng2024maelm}). BarcodeMAE is designed to address the MLM inefficiency with BERT-style transformer models for biodiversity analysis using DNA barcodes. This approach eliminates \texttt{[MASK]} tokens during encoding, thereby mitigating the distribution shift between pretraining and inference. Computation and parameters needed to predict values for \texttt{[MASK]} tokens is isolated to the decoder block, which is discarded after pretraining and not called at inference time. We empirically show that this mismatch is particularly detrimental in genomic pipelines where models are used for feature extraction without fine-tuning. 
To evaluate our model, we conduct self-supervised pretraining on the BIOSCAN-5M dataset \citep{gharaee2024bioscan5m}, which comprises over 2 million unique DNA barcodes. 
Our model outperforms existing foundation models in genus-level classification, surpassing a comparable encoder-only architecture by over 10 percentage points. Although it does not achieve the highest performance in BIN reconstruction, BarcodeMAE demonstrates superior average performance across evaluation tasks.




\section{Method \label{sec:method}}

In this section, we first describe the BIOSCAN-5M dataset and its partitioning scheme. Next, we introduce BarcodeMAE, our proposed model that adapts the masked auto-encoder architecture to address the representational limitations of masking approaches for DNA foundation models.

\subsection{Data \label{sec:data}}
Our analysis utilizes the BIOSCAN-5M dataset\footnote{The BIOSCAN-5M dataset contains 5.15\,M arthropod records, each with associated an image and DNA barcode sequence. Although the images are different for each record, the same barcode can occur across multiple records, hence there are fewer than 5\,M unique barcodes.
}, a comprehensive collection of 2.4\,M unique DNA barcodes organized into three distinct partitions: {\it (i) Pretrain}: Contains 2.28\,M unique DNA barcodes from unclassified specimens, used for self-supervised pretraining. {\it (ii) Seen}: Encompasses DNA barcodes with validated scientific species names, split into training (118\,k barcodes), validation (6.6\,k barcodes), and test (18.4\,k barcodes) subsets for closed-world evaluation tasks. {\it (iii) Unseen}: Contains novel species with reliable placeholder taxonomic labels, distributed across reference key (12.2\,k barcodes), validation (2.4\,k barcodes), and test (3.4\,k barcodes) subsets for open-world evaluation tasks.
For each sample in \textit{unseen}, its species does not appear in \textit{seen}, but its genus does appear.
%
This structure enables the evaluation of both closed-world classification and open-world species identification capabilities.

\subsection{BarcodeMAE: A masked auto-encoding model for DNA barcode sequences\label{sec:mask_strategies}}
In this section, we present BarcodeMAE, the encoder-decoder architecture for DNA sequence modelling. We describe the core architectural design and masking strategy that addresses the distribution shift between pretraining and inference, followed by the implementation specifications: tokenization, positional embeddings, and training procedures.

\subsubsection{Encoder-decoder architecture with modified masking}

To address the representational inefficiency in DNA sequence modelling, we adapt the MAE-LM approach \citep{meng2024maelm} for genomic applications.
In training using masked language modelling objectives, part of the encoder's capacity must be allocated to processing \texttt{[MASK]} tokens, which potentially limits the model's overall representational capacity to encode real tokens. The MAE-LM architecture effectively mitigates this limitation by using a bidirectional encoder and shallow bidirectional decoder, where the masked tokens are only presented to the decoder.


The encoder operates on nucleotide sequences with masked-out tokens removed entirely. Given a DNA sequence $\mathbf{x} = [x_1, \ldots, [\texttt{MASK}]_i, \ldots, x_n]$ and the set of masked positions $\mathcal{M}$, the encoder processes only nucleotide tokens. The encoder's input sequence $\mathbf{H}^0$ is composed of token embeddings $\mathbf{e}_{x_i}$ and positional embeddings $\mathbf{p}_i$ for non-masked positions:
\begin{equation}
\mathbf{H}^0 = \{h_i^0\}_{i\notin\mathcal{M}}, \quad h_i^0 = \mathbf{e}_{x_i} + \mathbf{p}_i
\end{equation}
The decoder then processes sequences containing both masked and unmasked positions, explicitly incorporating the \texttt{[MASK]} token in its input. The decoder's input sequence $\hat{\mathbf{H}}^0$ is constructed as:
\begin{equation}
\hat{\mathbf{H}}^0 = \{\hat{h}_i^0\}_{1\leq i\leq n}, \quad
\hat{h}_i^0 = \begin{cases}
\mathbf{e}_{[\text{MASK}]} + \mathbf{p}_i & i \in \mathcal{M} \\
h_L^i + \mathbf{p}_i & i \notin \mathcal{M}
\end{cases}
\end{equation}
where $h_L^i$ represents the final layer output from the encoder for non-masked positions and ${e}_{[\text{MASK}]}$ is the token embedding for the \texttt{[MASK]} token.

This approach prevents the encoder from learning specific embeddings for the \texttt{[MASK]} token, ensuring the decoder's representational capacity is not devoted to encoding this special token. Consequently, the encoder's representations remain unaffected by the \texttt{[MASK]} token and will use the full representational capacity to learn meaningful patterns from the nucleotide sequences. During downstream tasks, only the encoder is utilized, effectively isolating any potential limitations or inefficiencies related to the \texttt{[MASK]} tokens.

\subsubsection{Model implementation}

\begin{figure}[!h]
    \centering
    \includegraphics[width=1\linewidth]{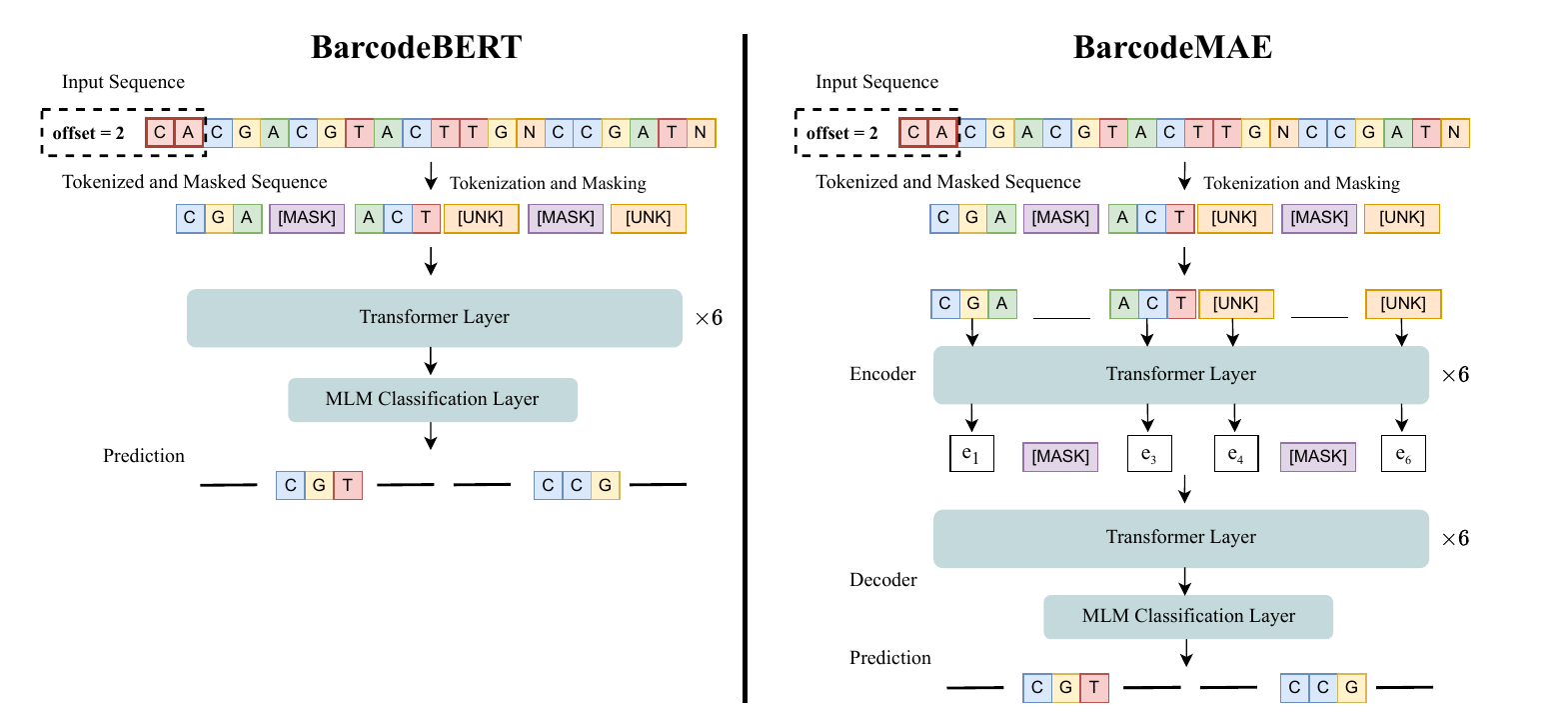}
    \caption{\textit{Comparison of pretraining processes for BarcodeBERT (left) and BarcodeMAE (right)}. 
    BarcodeBERT uses an encoder-only transformer architecture with direct masking. BarcodeMAE processes DNA barcode sequences through a transformer encoder-decoder architecture. The masking strategy differs from other foundation models by excluding the \texttt{[MASK]} token from the encoder input, requiring the decoder to predict masked sequences. After pretraining, the decoder is discarded and only the encoder is used for downstream tasks.}
    \label{fig:BarcodeMAE}
\end{figure}

BarcodeMAE uses a transformer architecture to implement the MAE-LM framework for DNA barcodes. It is trained using masked language modelling objectives. The architecture consists of a symmetrical design: an encoder and decoder, each comprising 6 transformer layers with 6 attention heads. Both components maintain a consistent hidden dimension of 768 units to ensure uniform representation capacity throughout the network. 
To obtain an embedding of the entire DNA barcode, the model employs global average pooling across the sequence of 768-dimensional output vectors, excluding padding and special tokens. \autoref{fig:BarcodeMAE} illustrates the architectural differences between BarcodeBERT, an encoder-only foundation model, and BarcodeMAE, an encoder-decoder model.

For DNA sequence processing, we use non-overlapping $k$-mer tokenization with a vocabulary size of $4^k + 2$, including the \texttt{[UNK]} and \texttt{[MASK]} special tokens. To handle frame-shift sensitivity, we incorporate the data augmentation strategy proposed in BarcodeBERT \citep{arias2023barcodebert}, where sequences are randomly offset before tokenization. Based on previous studies \citep{arias2023barcodebert, NucleotideTransformerDalla-Torre2023} showing optimal performance with $k$ values of 4 or 6, we evaluate our model using both of these $k$-mer lengths.

In this model, the encoder processes DNA sequences without \texttt{[MASK]} tokens, requiring a modified positional encoding scheme. Our implementation preserves sequence order by skipping masked position indices during encoding. This design maintains the relative positions of unmasked tokens from the original sequence, enabling spatial relationship modelling in DNA sequences.

We implement our model using PyTorch and the Hugging Face Transformers library. Our model is trained using masked token prediction with a 50\% token masking strategy.  To optimize the cross-entropy loss of masked tokens, we use AdamW \citep{AdamW} with a weight decay coefficient of $1\times10^{-5}$ and a OneCycle scheduler with a maximum learning rate of $1 \times 10^{-4}$.

\section{Experiments}

In this section, we describe both closed-world and open-world evaluation tasks designed to assess different aspects of the model's performance. Additionally, we present comparative results against current state-of-the-art baselines and conclude with an ablation study examining the impact of $k$-mer length and the number of layers.

\subsection{Evaluation framework}
We evaluate our model through two self-supervised learning (SSL) tasks: a closed-world task assessing generalization to new species within known genera, and an open-world task evaluating the model's ability to handle unseen taxonomic groups.

\textbf{Closed-World Task: 1-NN Probing.} To evaluate model generalization to new species within known genera, we perform genus-level 1-NN classification using cosine similarity. We use the training subset of the \textit{Seen} partition as the reference set and the \textit{Unseen} partition as the query set. This task, while involving unseen species, operates within the closed-world setting as it evaluates performance on known genera from the training taxonomy.

\textbf{Open-World Task: BIN Reconstruction.} To assess the model's ability to identify novel species and capture taxonomic relationships, we implement a Barcode Index Number (BIN) reconstruction task. We merge the test subset from the \textit{Seen} partition with the test subset of \textit{Unseen} partition and employ zero-shot clustering on embeddings generated without fine-tuning \citep{zsc-Lowe-2024}. This evaluation is particularly crucial for understanding the model's capability to group sequences from rare or previously unclassified species based on shared biological features.

\subsection{Results}
\label{s:results}

We compared BarcodeMAE against a comprehensive set of baselines, four encoder-only transformer-based models, DNABERT-2 \citep{zhou2023dnabert2}, DNABERT-S \citep{zhou2024dnaberts}, Nucleotide Transformer \citep{NucleotideTransformerDalla-Torre2023}, all trained on non-barcode data, and BarcodeBERT \citep{arias2023barcodebert}, trained on DNA barcodes. Since BarcodeBERT is a 4-layer encoder-only model and BarcodeMAE uses 6 encoder layers we pretrained a 6-layer model on the BIOSCAN-5M to ensure the fair comparison. We also implemented a baseline that uses an encoder-decoder architecture whilst maintaining the standard masking (BarcodeMAE w/MASK). This serves as a controlled baseline to isolate the impact of architectural choices from masking strategies. Note that, even though BarcodeMAE is conceptually an encoder-decoder model, for both BarcodeMAE and BarcodeMAE w/MASK, we discard the decoder component at inference time, using only the pretrained encoder for downstream tasks.

As shown in \autoref{tab:main_results}, BarcodeMAE achieves state-of-the-art performance in genus-level classification with 69.0\% accuracy, significantly outperforming the previous best baseline, BarcodeBERT, by over 10\%. This strong performance in the 1-NN probing task suggests that BarcodeMAE develops more effective representations of the taxonomic hierarchy, particularly in closed-world scenarios where the genera are known but the species are unseen. Notably, even our BarcodeMAE w/MASK baseline model outperforms existing approaches, demonstrating that decoupling the encoder and decoder alone contributes to improving representation learning in DNA barcode sequences, independent of masking strategy optimizations.

For BIN reconstruction using ZSC, DNABERT-S achieves the highest AMI score of 87.7\%, potentially due to its diverse pretraining dataset that aligns well with the clustering objective \citep{zhou2024dnaberts}. Notably, BarcodeMAE reaches comparable performance with an AMI of 80.3\%, outperforming models like DNABERT-2 and BarcodeBERT. To assess the performance across closed and open-world tasks, we calculated the harmonic mean between genus-level accuracy and BIN reconstruction AMI. BarcodeMAE achieves the highest harmonic mean of 74.2\%, outperforming all other baselines. This balanced metric highlights BarcodeMAE's robust performance across both genus-level classification and BIN reconstruction tasks.

\begin{table}[!h]
\caption{%
\textit{Performance comparison of DNA foundation models on  BIOSCAN-5M.} We evaluate on two key tasks: genus-level accuracy for 1-NN probing of unseen species and BIN reconstruction AMI using ZSC. The harmonic mean between these metrics provides a balanced assessment of each model's performance across both tasks. The models are divided into two groups: encoder-only transformer-based DNA foundation models, and our proposed model, BarcodeMAE. The \best{best} results are indicated in {bold}, and \secbest{second best} {underlined}. BarcodeMAE achieves the highest harmonic mean of 74.2\%, demonstrating superior balanced performance across closed and open-world tasks.}
\label{tab:main_results}
\begin{center}
\resizebox{\textwidth}{!}{%
\footnotesize
\begin{tabular}{@{}lllrrr@{}}
\toprule
& & &\makecell{\bf Genus-level acc (\%) \\ \bf of unseen species} & \makecell{\bf BIN clustering \\ \bf AMI (\%)} & \multirow[b]{2}{*}[-1ex]{\makecell{\bf Harmonic\\\bf Mean}} \\
\cmidrule(l){4-4}
\cmidrule(l){5-5}
{\bf Architecture}&{\bf  SSL Pretraining}&{\bf Model}   & \bf 1-NN probe & \bf ZSC probe \\
\midrule
Encoder-only   & Multi-species DNA & DNABERT-2              &         18.0 &         77.0 &         29.2 \\
               & Multi-species DNA & DNABERT-S              &         17.7 &   \best{87.7}&         29.5 \\
               & Multi-species DNA & Nucleotide Transformer &         21.7 &         37.3 &         27.4 \\
               & BIOSCAN-5M        & BarcodeBERT            &         58.3 &         79.3 &         67.2 \\
             
\midrule
Encoder-decoder& BIOSCAN-5M        & BarcodeMAE w/MASK      &\secbest{65.4}&\secbest{80.6}&\secbest{72.2}\\
               & BIOSCAN-5M        & BarcodeMAE             &   \best{69.0}&         80.3 &   \best{74.2}\\ 
\bottomrule
\end{tabular}
}
\end{center}
\end{table}

\begin{figure}[h]
    \centering \includegraphics[width=\linewidth]{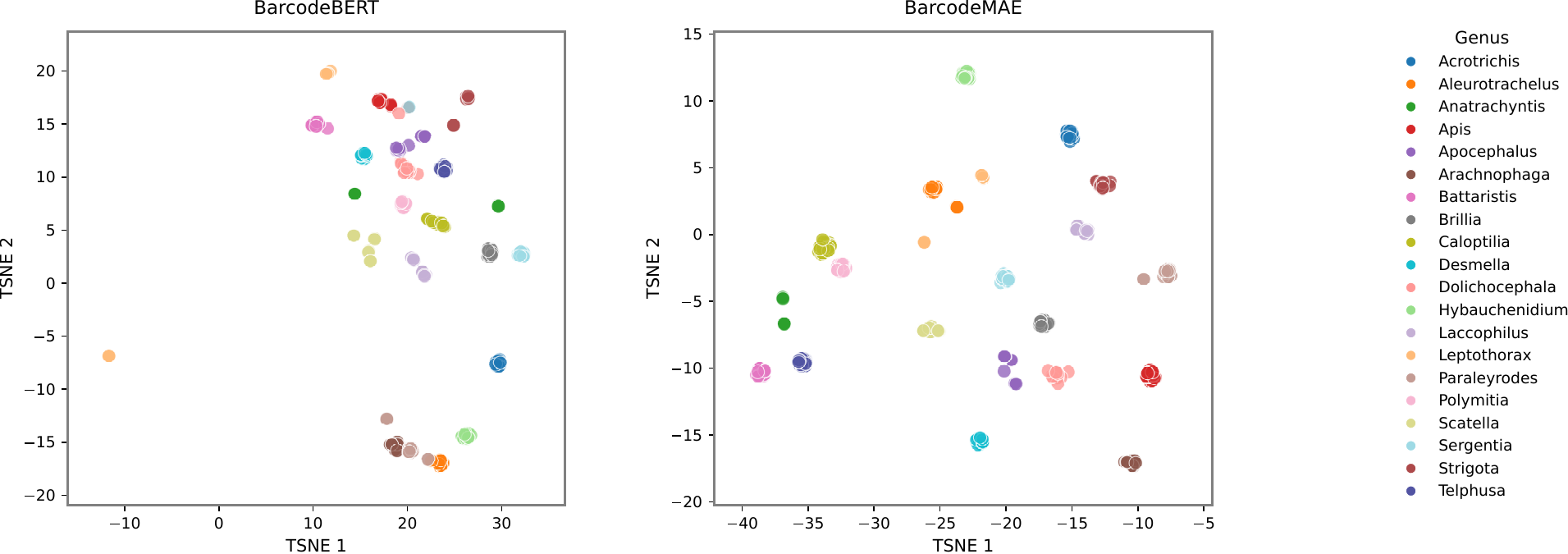}
    \caption{%
    \textit{t-SNE visualization of DNA barcode embeddings} from BarcodeBERT (left) and BarcodeMAE (right) for 20 randomly selected underrepresented genera. Each point represents a DNA barcode sequence, and colours indicate different genera. BarcodeMAE shows more distinct and well-separated clusters, suggesting better discrimination between genera compared to BarcodeBERT.}
    \label{fig:emb}
\end{figure}

To further validate the effectiveness of the model on underrepresented taxa, \autoref{fig:emb} visualizes the embeddings for 20 randomly sampled genera with fewer than 50 sequences in the dataset.  The embeddings from BarcodeMAE and the second best-performing model, BarcodeBERT, are projected to two dimensions using t-SNE \citep{vanDerMaaten2008}. The visualization shows that BarcodeMAE produces more cohesive and well-separated clusters compared to BarcodeBERT, indicating its ability to learn more discriminative embeddings even for genera with limited samples. 


\subsection{Ablation study}
We conducted an ablation study to analyze the impact of different architectural configurations on BarcodeMAE model performance, focusing on two key parameters: $k$-mer size and the number of layers and attention heads of the encoder-decoder. As shown in \autoref{tab:maelm-ablation}, we systematically varied the number of layers in both the encoder and decoder. The notation ``enc:L-H dec:M-J" indicates an encoder with L layers and H attention heads and a decoder with M layers and J attention heads. For each architecture, we evaluated both a $k$-mer size of $k=4$ and $k=6$.

\begin{table}[h]
\caption{%
\textit{Impact of $k$-mer size and model architecture on genus-level classification accuracy.} Architecture notation ``enc:L-H dec:M-J" indicates L layers and H attention heads in the encoder, and M layers and J heads in the decoder. Results are shown for $k=4$ and $k=6$, with \best{best} performance per $k$-mer size in bold, and \secbest{second best} performance underlined. }
\label{tab:maelm-ablation}
\begin{center}
\small
\begin{tabular}{cccccc} 
\toprule
& \multicolumn{5}{c}{Genus-level acc (\%) of} \\
 & \multicolumn{5}{c}{unseen species with $1$-NN probe} \\
\cmidrule(l){2-6}
\multicolumn{1}{c}{$k$-mer size} & \makecell{enc:4-4 \\ dec:2-2} & \makecell{enc:4-4 \\ dec:4-4} & \makecell{enc:6-6 \\ dec:2-2} & 


\makecell{enc:6-6 \\ dec:4-4} &\makecell{enc:6-6 \\ dec:6-6} \\
\midrule
4 & 64.1 & \best{68.4} & 65.0 & \secbest{66.1} & 65.3 \\
6 & 60.5 & 64.9 & 64.0  & \secbest{67.1} & \best{69.0} \\
\bottomrule
\end{tabular}
\end{center}
\end{table}

Our experiments demonstrate that the best performance is achieved with balanced encoder and decoder architectures (enc:6-6 dec:6-6), achieving 69.0\% accuracy for $k=6$. This contradicts traditional NLP approaches where shallower decoders are preferred \citep{meng2024maelm}. The improved performance with deeper decoders indicates that DNA sequence modelling requires more complex feature reconstruction capabilities. This finding provides evidence that effective DNA language models need architectures specifically designed for genomic data rather than direct adaptations from NLP.\looseness=-1

\subsection{Empirical evidence of representational deficiency in DNA foundation models}

In this section, we investigate the effects of \texttt{[MASK]} token embeddings across both the encoder-only foundation model, BarcodeBERT, and our proposed encoder-decoder model, BarcodeMAE.
To understand how the presence of \texttt{[MASK]} tokens impact taxonomic information during inference, we conducted two experiments using the genus-level 1-NN classification task from \autoref{s:results}. First, we replaced different portions of input sequences with \texttt{[MASK]} tokens, varying the masking ratio from 0.1 to 0.9, and evaluated the performance of the pretrained BarcodeBERT model (for which the encoder saw \texttt{[MASK]} tokens during training). Second, we performed a comparative analysis where instead of substituting the dropped tokens with \texttt{[MASK]}, we instead, removed them entirely. This version was performed for both BarcodeMAE and BarcodeBERT.
This allows us to study the impact of removing portions of the sequence on both models, and the effect of the \texttt{[MASK]} token versus token deletion on model performance.

\begin{figure}[tbh]
    \centering
    \includegraphics[width=0.65\linewidth]{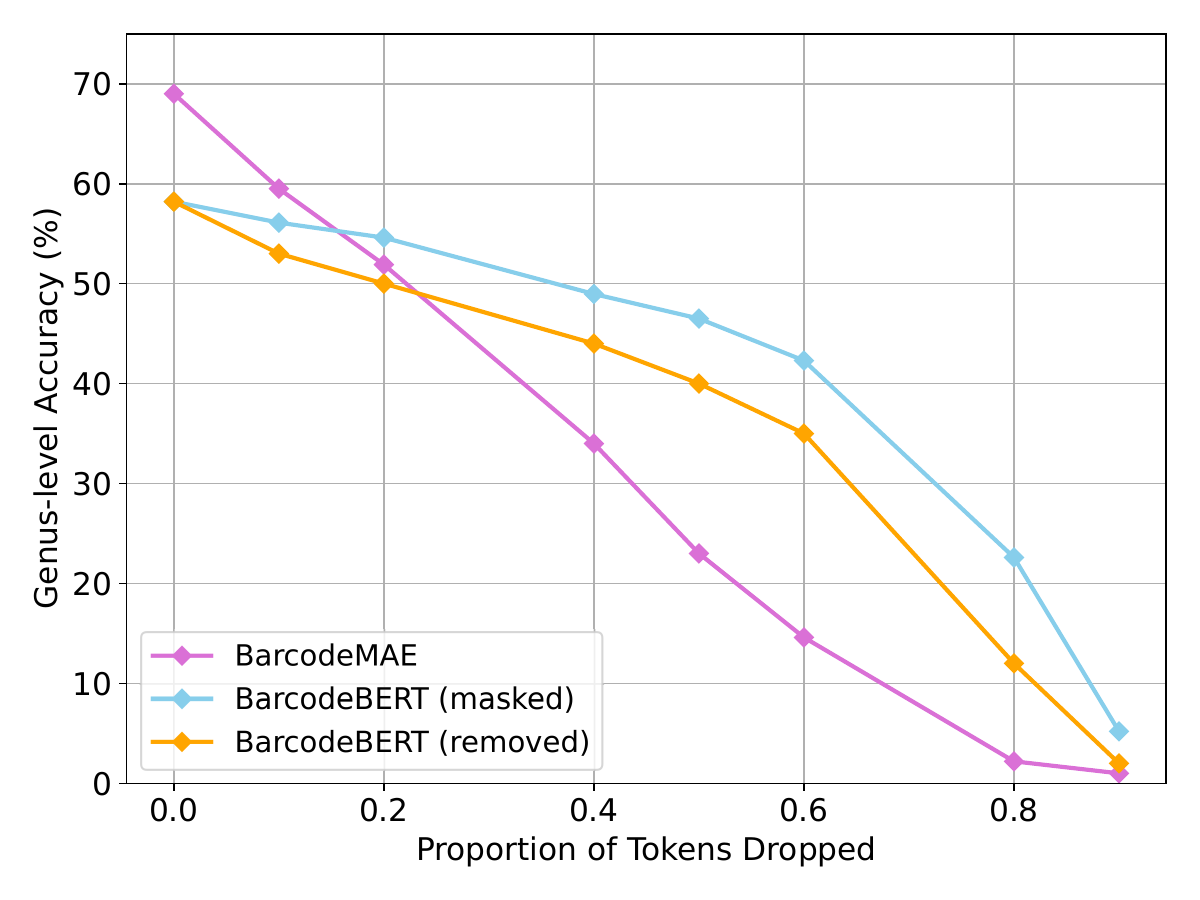}
    \caption{\textit{Impact of masking and token deletion on genus-level classification accuracy.} While BarcodeBERT shows stability at higher drop rates, the practical inference scenario occurs at $x\!=\!0$ with no masking, where BarcodeMAE demonstrates superior performance. The robustness to masking or removing tokens shown by BarcodeBERT does not correspond to an improved real-world performance since these conditions are not encountered during inference.}
    \label{fig:emp_mask}
\end{figure}

As shown in \autoref{fig:emp_mask}, we find that BarcodeMAE outperforms BarcodeBERT in the expected downstream use-case where the whole input sequence is shown to the model.
The performance of BarcodeMAE decreases approximately linearly as input tokens are removed from the sequence, and begins to fall as soon as any tokens are removed.
Meanwhile, the BarcodeBERT model, in both masked and removed variants, demonstrates only a shallow decline in performance as tokens are dropped until reaching its training masking ratio of 50\%, after which the accuracy decreases much more rapidly.
These results demonstrate that BarcodeBERT is better able to operate on partially complete information, but can not integrate together all information in the sequence.
Given that the two training tasks are the same, it is surprising that BarcodeMAE does not match the performance of BarcodeBERT for partial sequences, and this suggests there may be potential for further performance gains.

Additionally, we find that BarcodeBERT performs better when dropped tokens are replaced with the \texttt{[MASK]} token instead of being removed completely from the input. The performance gap emerges immediately (+4\% at 10\% dropped) and increases to reach approximately +10\% at 80\% dropped.
Since the \texttt{[MASK]} tokens do not contain any information about the specimen's genus, the fact that the BarcodeBERT model performs better when they are present indicates it learned to use the computation associated with \texttt{[MASK]} tokens to better extract information from rest of the sequence.

These results empirically demonstrate the distribution shift challenge inherent in masked language modelling, as the model develops dependency on \texttt{[MASK]} tokens during training that are absent during inference. The contrasting behaviour of BarcodeMAE, which learns representations solely from observed nucleotides, suggests its architecture may better align with inference-time conditions, where the \texttt{[MASK]} token is not present.

\section{Conclusion}

We introduced BarcodeMAE, an encoder-decoder architecture that mitigates the fundamental limitations of masked language modelling in DNA barcode sequence analysis. By eliminating \texttt{[MASK]} tokens during encoding, BarcodeMAE reduces the distribution shift between pretraining and inference, significantly enhancing performance over existing DNA foundation models. Notably, it achieves over a 10\% improvement in genus-level classification accuracy on the BIOSCAN-5M dataset compared to the previous state-of-the-art, BarcodeBERT.

While BarcodeMAE does not have the best performance in BIN reconstruction, it achieves the highest harmonic mean across the evaluation tasks, demonstrating a robust performance between closed-world and open-world settings. Our ablation studies reveal that, unlike NLP models that favour shallow decoders, DNA sequence modelling benefits from balanced encoder-decoder architectures, underscoring the need for domain-specific architectural designs.

These findings highlight the critical impact of \texttt{[MASK]} token distribution shifts on foundation model effectiveness, particularly in genomic applications where models are used for feature extraction without fine-tuning. The superior performance of BarcodeMAE across diverse evaluation scenarios validates its architectural approach to addressing masking inefficiencies in genomic foundation models.

\bibliography{main.bib}
\bibliographystyle{iclr2025_conference}
\newpage
\appendix
\section*{Appendices}

\section{Comparison with causal models \label{appendix_1}}

To compare our model with recently developed causal models for DNA sequence analysis, we conducted additional experiments comparing BarcodeMAE with several state-of-the-art models, such as HyenaDNA-tiny and Caduceus-PS-1k which are trained on non-barcode data and BarcodeMamba which is trained on DNA barcode data. For a fair comparison, since BarcodeMamba was trained on the CANADA-1.5M dataset \citep{Hebert2016}, we trained BarcodeMAE on the same dataset and evaluated all models on BIOSCAN-5M.

As shown in \autoref{tab:appendix_results}, While causal models show strong performance in BIN clustering, with HyenaDNA-tiny achieving 85.0\% AMI, they underperform in the genus-level classification of unseen species. BarcodeMamba, specifically trained on DNA barcodes, achieves the highest balanced performance among state space models with 36.3\% genus-level accuracy and 82.7\% BIN clustering AMI, resulting in a harmonic mean of 50.5\%. 

The encoder-only architecture, BarcodeBERT, demonstrates enhanced genus-level classification through 1-NN probing compared to causal models, achieving 40.9\% accuracy. BarcodeMAE surpasses all competing models with 51.2\% genus-level classification accuracy and a harmonic mean of 63.2\%, indicating superior balanced performance across metrics. One interesting finding of these results is that BarcodeMAE surpasses models pre-trained on BIOSCAN-5M by 3\% in the ZSC bin reconstruction task, despite being trained on the smaller CANADA-1.5M dataset.

\begin{table}[!h]
\caption{%
\textit{Performance comparison of DNA foundation models on  BIOSCAN-5M.} We evaluate on two key tasks: genus-level accuracy for 1-NN probing of unseen species and BIN reconstruction AMI using ZSC. The harmonic mean between these metrics provides a balanced assessment of each model's performance across both tasks. The models are divided into three groups: transformer-based DNA foundation models, state space models, and our proposed model, BarcodeMAE. The \best{best} results are indicated in bold, and \secbest{second best} are underlined.}
\label{tab:appendix_results}
\begin{center}
\resizebox{\columnwidth}{!}{
\footnotesize
\begin{tabular}{@{}lllrrr@{}} 
\toprule
& & &\makecell{\bf Genus-level acc (\%) \\ \bf of unseen species} & \makecell{\bf BIN clustering \\ \bf AMI (\%)} & \multirow[b]{2}{*}[-1ex]{\makecell{\bf Harmonic\\\bf Mean}} \\
\cmidrule(l){4-4}
\cmidrule(l){5-5}
{\bf Architecture} & {\bf  SSL Pretraining} & {\bf Model}       & \bf 1-NN probe & \bf ZSC probe \\
\midrule
State space  & Human genome &HyenaDNA-tiny            &        19.3 &      \textbf{85.0} &      31.4 \\
             & Human genome &Caduceus-PS-1k           &         9.1 &      65.4 &      15.9 \\
             & CANADA-1.5M &BarcodeMamba             &        36.3 &      82.7 &      50.5 \\
\midrule
Encoder-only  & CANADA-1.5M &BarcodeBERT               &         40.9&       73.4&      52.5\\
\midrule
Encoder-decoder & CANADA-1.5M & BarcodeMAE w/MASK     & \secbest{49.4}&\secbest{83.8}&\secbest{62.2}\\
                & CANADA-1.5M & BarcodeMAE            & \best{51.2}&      82.4 &\best{63.2}\\ 
\bottomrule
\end{tabular}
}
\end{center}
\end{table}

\section{Baseline models}
For evaluation, we utilized the respective Pretrained models from Huggingface ModelHub, specifically:

\begin{itemize}
\item DNABERT-2: \href{https://huggingface.co/zhihan1996/DNABERT-2-117M}{zhihan1996/DNABERT-2-117M}
\item DNABERT-S: \href{https://huggingface.co/zhihan1996/DNABERT-S}{zhihan1996/DNABERT-S}
\item NT: \href{https://huggingface.co/InstaDeepAI/nucleotide-transformer-v2-50m-multi-species}{InstaDeepAI/nucleotide-transformer-v2-50m-multi-species}
\item HyenaDNA: \href{https://huggingface.co/LongSafari/hyenadna-tiny-1k-seqlen}{LongSafari/hyenadna-tiny-1k-seqlen}
\item BarcodeBERT: \href{https://huggingface.co/bioscan-ml/BarcodeBERT}{bioscan-ml/BarcodeBERT}
\item Caduceous: \href{https://huggingface.co/kuleshov-group/caduceus-ps_seqlen-1k_d_model-256_n_layer-4_lr-8e-3}{kuleshov-group/caduceus-ps\_seqlen-1k\_d\_model-256\_n\_layer-4\_lr-8e-3}
\item BarcodeMamba: \href{https://huggingface.co/bioscan-ml/BarcodeMamba}{bioscan-ml/BarcodeMamba-dim384-layer2-char}
\end{itemize}

\subsection{Pretraining}
BarcodeBERT and BarcodeMAE were pretrained for 35 epochs using the AdamW optimizer \cite{AdamW} with a learning rate of $2\times10^{-4}$, a batch size of 128, and a OneCycle learning rate scheduler. The pretraining process utilized four NVIDIA V100 GPUs and required approximately 36 hours to complete for each experiment executed. To examine the impact of pretraining, we also trained a model from scratch on the training subset of the \textit{Seen} partition without any pretraining. 

\subsection{Zero-shot clustering}

We evaluated the models’ ability to group sequences without supervision using a modified version of the framework from \iftoggle{arxiv}{\citet{zsc-Lowe-2024}}{Lowe et. al. \cite{zsc-Lowe-2024}}. Embeddings were extracted from the pretrained encoders and reduced to 50 dimensions using UMAP \citep{umap} to enhance computational efficiency while preserving data structure. These reduced embeddings were clustered with Agglomerative Clustering (cosine distance, Ward’s linkage), using the number of true species as the target number of clusters. Clustering performance was assessed with adjusted mutual information (AMI) to measure alignment with ground-truth labels.

\end{document}